\documentclass[conference,dvipsnames]{IEEEtran}
\IEEEoverridecommandlockouts
\usepackage{cite}
\usepackage{amsmath,amssymb,amsfonts}
\usepackage{algorithmic}
\usepackage{graphicx}
\usepackage{textcomp}
\usepackage{xcolor}

\usepackage{float}
\usepackage{subfig}

\usepackage{epsfig}
\usepackage{multirow}
\usepackage{adjustbox}
\usepackage{marvosym}

\usepackage{pifont}

\definecolor{color3}{rgb}{0.95,0.95,0.95}
\definecolor{color4}{rgb}{0.96,0.96,0.86}
\definecolor{color5}{rgb}{0.90,0.90,0.90}

\definecolor{cvprblue}{rgb}{0.21,0.49,0.74}
\usepackage[pagebackref,breaklinks,colorlinks,allcolors=cvprblue]{hyperref}

\usepackage{xcolor}

\usepackage[abs]{overpic}
\usepackage{varwidth}
\usepackage{wrapfig}
\usepackage{diagbox}
\usepackage{extpfeil}
\usepackage{makecell}
\usepackage{colortbl}
\usepackage{float}
\usepackage{wrapfig}
\usepackage{booktabs}

\def\BibTeX{{\rm B\kern-.05em{\sc i\kern-.025em b}\kern-.08em
    T\kern-.1667em\lower.7ex\hbox{E}\kern-.125emX}}
\begin{document}

\title{Learning Physics-Informed Color-Aware Transforms for Low-Light Image Enhancement}

\author{\IEEEauthorblockN{Xingxing Yang}
\IEEEauthorblockA{\textit{Department of Computer Science} \\
\textit{Hong Kong Baptist University}\\
Hong Kong SAR, China \\
csxxyang@comp.hkbu.edu.hk}
\and
\IEEEauthorblockN{Jie Chen$^{\ast}$\thanks{*Corresponding author}}
\IEEEauthorblockA{\textit{Department of Computer Science} \\
\textit{Hong Kong Baptist University}\\
Hong Kong SAR, China \\
chenjie@comp.hkbu.edu.hk}
\and
\IEEEauthorblockN{Zaifeng Yang}
\IEEEauthorblockA{\textit{Institute of High Performance Computing} \\
\textit{Agency for Science, Technology and Research}\\
Singapore \\
yang\_zaifeng@ihpc.a-star.edu.sg}
}


\maketitle

\begin{abstract}
Image decomposition offers deep insights into the imaging factors of visual data and significantly enhances various advanced computer vision tasks. In this work, we introduce a novel approach to low-light image enhancement based on decomposed physics-informed priors. Existing methods that directly map low-light to normal-light images in the sRGB color space suffer from inconsistent color predictions and high sensitivity to spectral power distribution (SPD) variations, resulting in unstable performance under diverse lighting conditions. To address these challenges, we introduce a Physics-informed Color-aware Transform (PiCat), a learning-based framework that converts low-light images from the sRGB color space into deep illumination-invariant descriptors via our proposed Color-aware Transform (CAT). This transformation enables robust handling of complex lighting and SPD variations. Complementing this, we propose the Content-Noise Decomposition Network (CNDN), which refines the descriptor distributions to better align with well-lit conditions by mitigating noise and other distortions, thereby effectively restoring content representations to low-light images. The CAT and the CNDN collectively act as a physical prior, guiding the transformation process from low-light to normal-light domains. 
Our proposed PiCat framework demonstrates superior performance compared to state-of-the-art methods across five benchmark datasets.

\end{abstract}

\begin{IEEEkeywords}
low-light, image decomposition, noise estimation, Lambertian assumption, dynamic filter
\end{IEEEkeywords}

\section{Introduction}
\label{sec:intro}

\begin{figure}[t]
    \centering
  { 
      \includegraphics[width=0.98\linewidth]{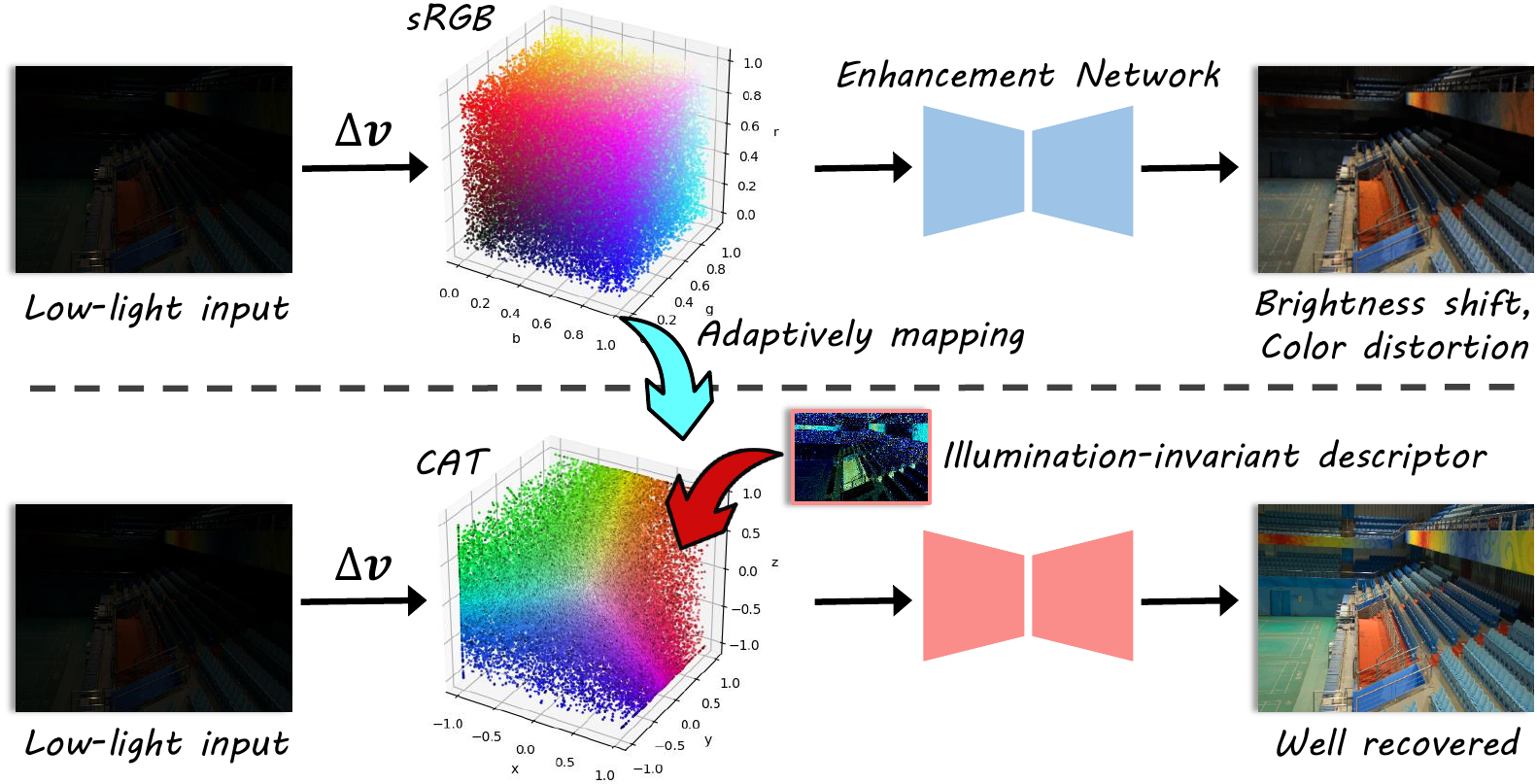}}
  \caption{Comparisons between our PiCat (bottom) and previous methods (top). The notation $\Delta\upsilon$ denotes tiny Gaussian noise in the frequency domain to simulate SPD perturbations. After enhancement, significant color distortions and brightness shifts can be observed in previous methods.}
    \label{fig:teaser} 
\vspace{-5pt}
\end{figure}

The pursuit of generating high-quality images from degraded sources has driven significant advancements in the field of Low-Light Image Enhancement (LLIE), which aims to improve the visibility of low-light observations while concurrently mitigating the influences of noise, color bias, and brightness shifts.
Most existing methods~\cite{cai2023retinexformer, snr_net, restormer} attempt to learn the mapping function between low- and normal-light images in the sRGB color space. However, this approach is characterized by a strong interdependence of brightness and color across its three channels, leading to unstable color predictions, brightness shifts, and increased sensitivity to variations in spectral power distribution (SPD).
As illustrated in Fig.~\ref{fig:teaser}, we simulate minor SPD disturbances by introducing Gaussian noise in the frequency domain to obtain new degraded testing samples.
The disturbances result in dramatic performance degradation in existing learning-based methods, suggesting that there is a mismatch between the sRGB color space and the degradation enhancement processing, leading to instability in both brightness estimation and color recovery.

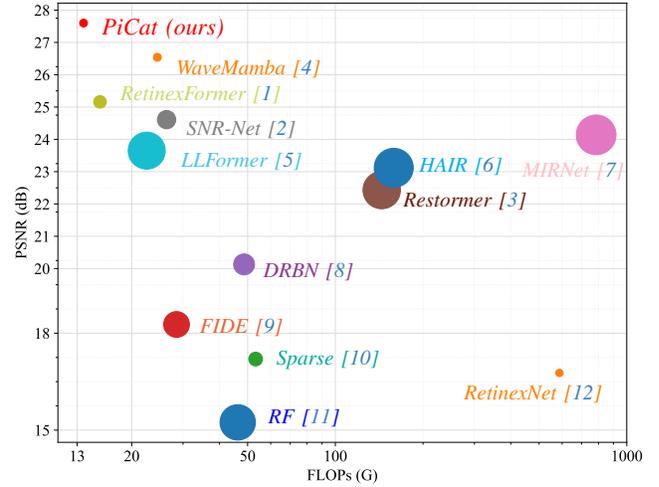
\begin{figure}

\begin{overpic}[width=0.46\textwidth]{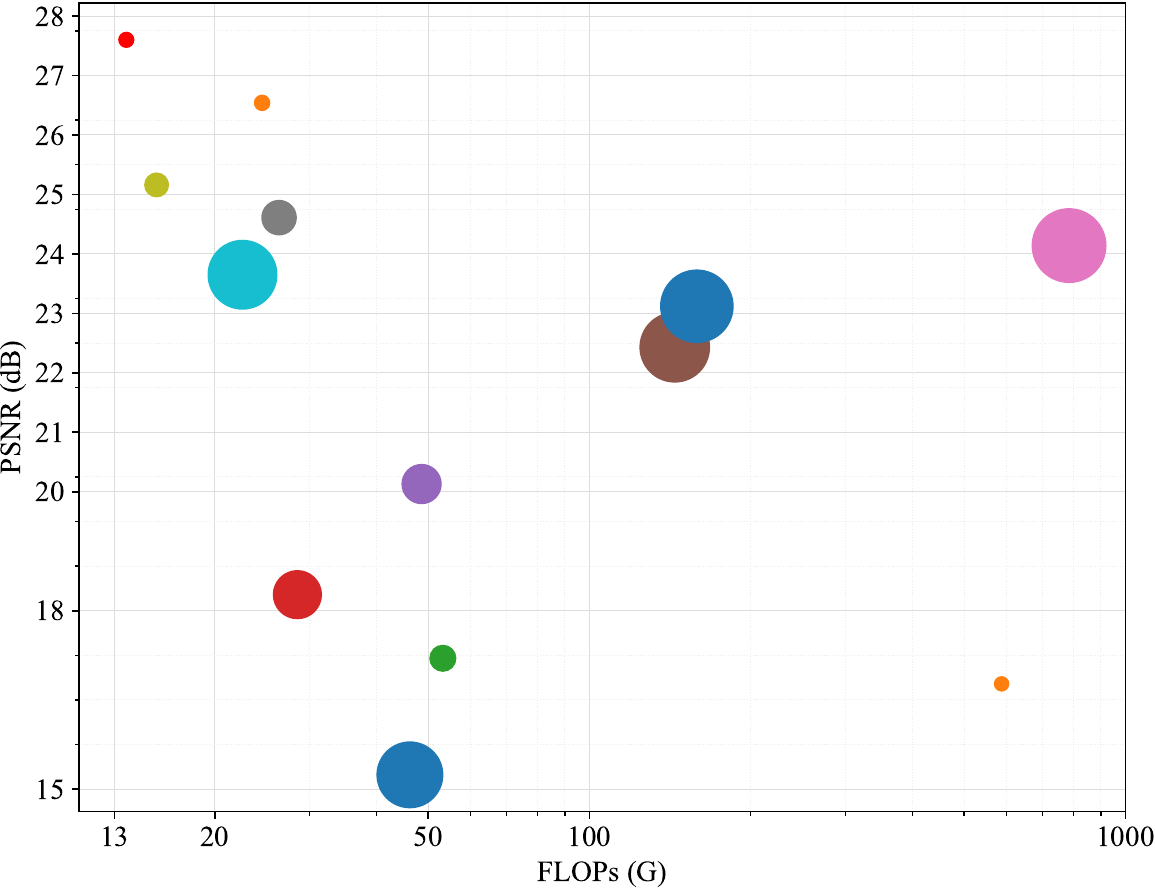}
\put(33,170){\color{red}\small \textit{PiCat~(ours)}} 
\put(61,155){\color{orange}\footnotesize \textit{WaveMamba~\cite{zou2024wave}}} 
\put(40,145){\color{SpringGreen}\footnotesize \textit{RetinexFormer~\cite{cai2023retinexformer}}} 
\put(65,132){\color{gray}\footnotesize \textit{SNR-Net~\cite{snr_net}}} 
\put(63,120){\color{SkyBlue}\footnotesize \textit{LLFormer~\cite{wang2023ultra}}} 
\put(153,118){\color{cyan}\footnotesize \textit{HAIR~\cite{cao2024hair}}} 
\put(192,116){\color{pink}\footnotesize \textit{MIRNet~\cite{mirnet}}} 
\put(147,105){\color{Sepia}\footnotesize \textit{Restormer~\cite{restormer}}} 
\put(94,78){\color{Fuchsia}\footnotesize \textit{DRBN~\cite{drbn}}} 
\put(70,57){\color{RedOrange}\footnotesize \textit{FIDE~\cite{fide}}} 
\put(99,45){\color{JungleGreen}\footnotesize \textit{Sparse~\cite{lol_v2}}} 
\put(96,23){\color{blue}\footnotesize \textit{RF~\cite{rf}}} 
\put(170,32){\color{orange}\footnotesize \textit{RetinexNet~\cite{retinex_net}}} 
\end{overpic}
\vspace{-3pt}
\caption{Performance comparisons. FLOPs(complexity) and PSNR (performance) are displayed along the horizontal and vertical axes, respectively. The size of the circles represents the parameter count (memory cost) using a logarithmic scale. Our PiCat outperforms SOTA methods by delivering the highest PSNR while significantly lowering computational costs.}
\label{fig:performance}
\vspace{-5pt}
\end{figure}

This leads us to a thought-provoking question: \textbf{\textit{Can we explore a new color transformation that is invariant to both illumination and SPD, thereby enabling more stable and robust enhancement?}}

To address this question, we need to summarize existing literature on LLIE and color transformation, building upon the contributions of earlier researchers to explore potential new solutions for illumination-/SPD-invariance descriptor extraction.
Traditional enhancement methods usually light up degraded images via physical models such as Gamma Correction~\cite{huang2012efficient} and Retinex Decomposition~\cite{land1971lightness}. These methods rely on fixed formulations that struggle to address the diverse and complex degradation scenarios, particularly under extremely dark conditions, which can result in color distortions and brightness shifts.
In response to these limitations, learning-based methods~\cite{xu2020learning, snr_net, retinex_net, zhou2022lednet} employ deep neural networks (\textit{e.g.}, CNNs) to learn a brute-force mapping function from the degraded image to its normal-light counterparts in the sRGB color space.
Nevertheless, directly learning the mapping function in the sRGB color space suffers from color distortion and sensitivity to SPD change. For instance, RetinexNet~\cite{retinex_net} proposes a CNN-based dual-branch network to predict reflectance and illumination maps in the sRGB color space according to Retinex theory. Although rough brightness estimation and color recovery can be achieved, conspicuous color distortion and artifacts occur in some extremely dark regions and different illumination conditions.

Meanwhile, some researchers have investigated the importance of color space and physical priors in various vision tasks~\cite{zhai2024multi, zhang2021better, gevers1999color}.
For example, MCFNet~\cite{zhai2024multi} introduces HSV color transform to tackle spectral translation through a series of escalating resolutions, progressively enriching images with color and texture fidelity.
However, the HSV color space is still sensitive to variations in spectral power distribution (SPD) and noise disturbances.
Additionally, Cross Colour Ratios (CCR)~\cite{gevers1999color} have been introduced to leverage illumination-invariant properties for object detection. Nonetheless, the CCR method relies on a fixed formulation based on several conditional assumptions, which restricts its ability to adapt to the diverse and complex illumination conditions in various real-world scenarios.

\begin{figure*}[t]
    \centering
  { 
      \includegraphics[width=0.97\linewidth]{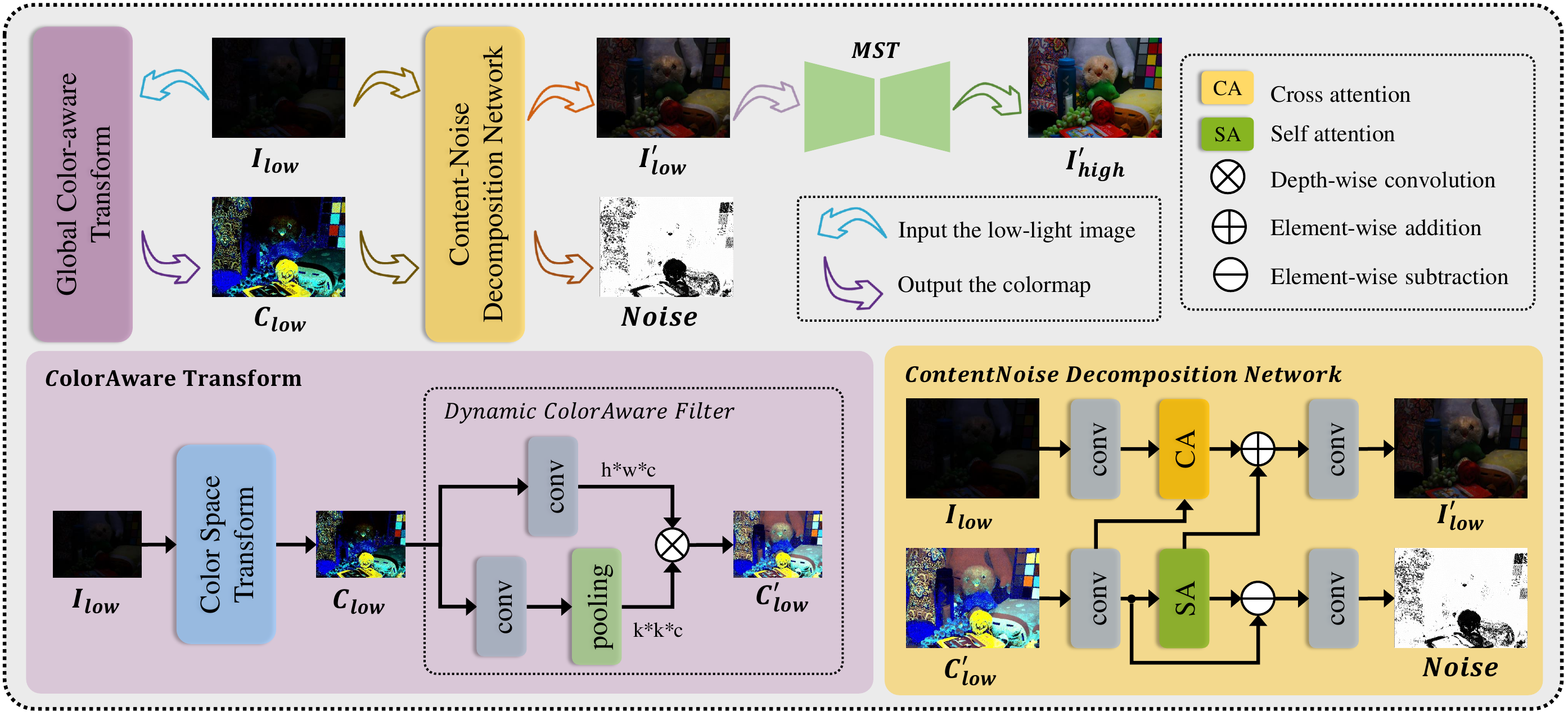}}
  \caption{Pipeline of \textbf{PiCat}. PiCat extracts the illumination-invariant descriptor via the Color-aware Transform (CAT). The Content-Noise Decomposition Network (CNDN) estimates and decomposes the noise distribution explicitly and transfers fine-grained content information from the illumination-invariant descriptor to the target low-light image. Eventually, MST~\cite{cai2022mask} recovers the final enhanced image.}
    \label{fig:pipeline} 
    \vspace{-5pt}
\end{figure*}

To address these issues, we propose a Physics-informed Color-aware Transform (PiCat), a learning-based framework that converts low-light images from the sRGB color space into deep illumination-invariant descriptors based on the Lambertian assumption, providing prior knowledge for downstream enhancement networks. 
The main difference between PiCat and existing transformation methods is summarized in Table~\ref{tab:property}.
Specifically, we first calculate the channel-wise color ratio of the three sRGB channels while incorporating a trainable color-sensitivity parameter.
This allows us to transform the degraded image into a new color space that exhibits illumination-invariant properties. 
Next, the illumination-invariant descriptor is processed by a dynamic color-aware filter to establish a robust photometric adaptation function capable of handling diverse lighting conditions.
Moreover, considering that low-light degradations are often accompanied by severe noise disturbances and color corruption, we introduce the Content-Noise Decomposition Network (CNDN) to explicitly estimate and decompose the noise distribution. 
This enables the transfer of noise-free, complementary content information from the illumination-invariant descriptor to the target low-light image. 
Finally, the contextually enhanced low-light image is processed by a downstream enhancement network for further improvement of visual effects.

Our contributions are summarized as follows:
\begin{itemize}
	\item We propose a learnable Physics-informed Color-aware Transform (PiCat) for LLIE, which exhibits illumination-invariant properties and is stable to spectral power distribution, enabling robust handling of complex lighting and SPD variations;
	\item We propose a Content-Noise Decomposition Network (CNDN) to estimate and decompose the noise distribution explicitly, which refines the descriptor distributions extracted by the Color-aware Transform (CAT) to better align with well-lit conditions by mitigating noise and other distortions, thereby effectively restoring content representations to low-light images;
	\item Our method, termed \textit{\textbf{PiCat}}, functions as a plug-and-play module that can be seamlessly integrated into existing models. The CAT and the CNDN collectively act as a physical prior, guiding the transformation process from low-light to normal-light domains. Extensive experiments demonstrate that PiCat outperforms SOTA methods while maintaining lower computational costs.
\end{itemize}

\begin{table}[t]\tiny
	\centering
        \caption{Photometric properties comparisons. Our CAT is fully invariant to both illumination and geometry and is trainable. SPD denotes the spectral power distribution.}\label{tab:property}
	\setlength\tabcolsep{2pt}
	\resizebox{0.4\textwidth}{!}{\hspace{-0.5mm}
		\begin{tabular}{l|ccc|cc}
			\toprule[0.10em]
			\multirow{2}{*}{Algorithm}     & \multicolumn{3}{c|}{Invariances}   & \multirow{2}{*}{Trainability}   \\ 
                 & Illumination & Geometry & SPD  \\ 
                 \midrule[0.05em]
			
			sRGB  & \ding{56}  &  \ding{56} &  \ding{56} &  \ding{56} \\
                HSV   &  \ding{56} &  \ding{52} &  \ding{56} &  \ding{56} \\
                CCR~\cite{gevers1999color}  & \ding{52}  & \ding{52}  &  \ding{52} & \ding{56}  \\
                CAT  & \ding{52}  & \ding{52}  &  \ding{52} & \ding{52}  \\

            \bottomrule[0.10em]
	\end{tabular}}
	\vspace{-3mm}
\end{table}

\section{Method}
\label{sec:method}

Based on the motivation to explore a new color transformation that maintains illumination-invariant properties and thus facilitates a stable LLIE, we introduce a new method called PiCat. PiCat employs a learnable Physics-informed Color-aware Transform (PiCat) to decompose the low-light image from the sRGB color space into an illumination-invariant descriptor based on the Lambertian assumption.
Given that low-light images often suffer from significant noise and color distortion, PiCat also integrates a Content-Noise Decomposition Network (CNDN) to accurately estimate and decompose the noise distribution. This allows for the transfer of a noise-free content representation from the illumination-invariant color descriptor to the low-light images, facilitating a stable enhancement.
Finally, the enhanced low-light image is fed into an off-the-shelf restoration network, MST~\cite{cai2022mask}, for further enhancement. The overview of our pipeline is illustrated in Fig.~\ref{fig:pipeline}.

\subsection{Physics-informed Color-aware Transform}
\textbf{Lambertian Assumption.} The image formation process can be modeled according to the body reflection term of the dichromatic reflection model \cite{gevers1999color} as follows:

\begin{equation}\label{eq1}
   I=m(\vec{n}, \vec{l}) \int_\lambda e(\lambda) \rho_b(\lambda) f(\lambda) \mathrm{d} \lambda.   
\end{equation}
Here, $I$ is the captured image; $m$ is an interactive function depending on light sources and the object geometry; $\vec{n}, \vec{l}$ denote the surface normal and the direction of the light source, respectively. $\lambda$ is the incoming light wavelength within the visible spectrum $\lambda$, while $e$ describes the SPD of the light source. Additionally, $\rho_b$ represents the intrinsic property (\textit{i.e.}, reflectance); $f$ corresponds to the spectral camera sensitivity function.
Assuming that the sensor sensitivities of the color camera operate in narrow bands, the imaging model can be discretized as follows:
\begin{equation}\label{eq2}
   C_{p}=m(\vec{n}, \vec{l}) e^{C_{p}}(\lambda) \rho^{C_{p}}(\lambda).  
\end{equation}
Here, for $C=\{R, G, B\}$, $C_{p}$ denotes a color channel $C$ for pixel $p$ in a sRGB image.
It is evident that all color values within the sRGB space are influenced by variations in illumination and SPD. In the context of LLIE, differences in exposure times or lighting conditions can lead to significant instability in the performance of existing methods.

\begin{table*}[t]
	\centering
        \caption{Quantitative comparisons on LOL (v1~\cite{retinex_net} and v2~\cite{lol_v2}), SID~\cite{sid}, and SMID~\cite{smid} datasets. The highest result is in \textcolor{red}{red} while the second highest result is in \textcolor{blue}{blue}. Our PigCat significantly outperforms SOTA algorithms.}\label{tab:quantitative}
	\setlength\tabcolsep{4pt}
	\resizebox{0.98\textwidth}{!}{\hspace{-0.5mm}
		\begin{tabular}{l|l|cc|cc|cc|cc|cc|cc}
			\toprule[0.15em]
			\multirow{2}{*}{Methods}      & \multirow{2}{*}{Venue} & \multicolumn{2}{c|}{Complexity}   & \multicolumn{2}{c|}{LOL-v1} & \multicolumn{2}{c|}{LOL-v2-real}  &\multicolumn{2}{c|}{LOL-v2-syn}   &\multicolumn{2}{c|}{SID} &\multicolumn{2}{c}{SMID}   \\ & & FLOPs (G) & Params (M) & PSNR & SSIM & PSNR & SSIM & PSNR & SSIM & PSNR & SSIM & PSNR & SSIM  \\ \midrule[0.15em]
			RF~\cite{rf}   & AAAI'20  &46.23 &21.54   & 15.23  &0.452  &14.05  &0.458   &15.97    &0.632   &16.44 &0.596 &23.11 &0.681 \\
			RetinexNet~\cite{retinex_net} & BMVC'18 & 587.47    & \textcolor{blue}{0.84} & 16.77    & 0.560   & 15.47        & 0.567  &17.13  &0.798  &16.48 &0.578 &22.83 &0.684\\
			Sparse~\cite{lol_v2}   & TIP'21  &53.26 &2.33   &17.20  &0.640  &20.06  &0.816   &22.05    &0.905     &18.68 &0.606 &25.48 &0.766 \\
			FIDE~\cite{fide}    & CVPR'20 &28.51 &8.62   & 18.27  &0.665  &16.85  &0.678   &15.20    &0.612    &18.34 &0.578 &24.42 &0.692  \\
			DRBN~\cite{drbn} & TIP'21 &48.61  &5.27   & 20.13   & 0.830    &20.29   & 0.831    & 23.22     & 0.927   &19.02 &0.577  &26.60 &0.781  \\
			Restormer~\cite{restormer} &CVPR'22   &144.25 &26.13   &22.43        &0.823        &19.94         &0.827         &21.41      &0.830         &22.27 &0.649 &26.97 &0.758  \\
			MIRNet~\cite{mirnet} &ECCV'20  &785 &31.76    &24.14   &0.830  &20.02   &0.820  &21.94  &0.876   &20.84 &0.605   &25.66 &0.762  \\    
			SNR-Net~\cite{snr_net} &CVPR'22  &26.35   &4.01  &24.61 &0.842  &21.48  &0.849 &24.14 &0.928 &22.87 &0.625 &28.49 &0.805 \\ 
			Retinexformer~\cite{cai2023retinexformer} &ICCV'23    &15.57  &1.61   &25.16    &0.845        &22.80   &0.840  &25.67 &0.930 &24.44 &0.680 &29.15 &0.815 \\ 
            LLFormer~\cite{wang2023ultra} & AAAI'23 & 22.52 & 24.55 & 23.65 & 0.816 & 20.06 & 0.792 & 24.04 & 0.909 & 23.17 & 0.641 & 28.31  & 0.796 \\
            HAIR~\cite{cao2024hair} & ArXiv'24 & 158.62 & 29.45 & 23.12 & 0.847 & 21.44 & 0.839 & 24.71 & 0.912 & 23.54 & 0.653 & 28.19  & 0.803 \\
            WaveMamba~\cite{zou2024wave} & MM'24 & 24.50 & 1.02 & 26.54 & \textcolor{blue}{0.883} & \textcolor{blue}{29.04} & \textcolor{blue}{0.901} & 29.11 & 0.940 & 25.12 & 0.703 & \textcolor{red}{30.13}  & \textcolor{blue}{0.885} \\
            \midrule[0.15em]
            \textbf{PiCat} \textit{(ours)} &~~~~~-- & \textcolor{red}{7.06}  & \textcolor{red}{0.49} & \textcolor{blue}{27.16} & 0.873 & 28.44 & 0.891 & \textcolor{blue}{29.39} & \textcolor{blue}{0.945} & \textcolor{blue}{25.50} & \textcolor{blue}{0.718} & 29.97  & 0.879 \\
            \textbf{PiCat-large} &~~~~~-- & \textcolor{blue}{13.67} & 0.97 & \textcolor{red}{27.51} & \textcolor{red}{0.890} & \textcolor{red}{29.31} & \textcolor{red}{0.917} & \textcolor{red}{30.07} & \textcolor{red}{0.955} & \textcolor{red}{25.87} & \textcolor{red}{0.732} & \textcolor{blue}{30.11}  & \textcolor{red}{0.887} \\
            \bottomrule[0.15em]
	\end{tabular}}
	\vspace{-3mm}
\end{table*}

\textbf{Color Space Transform.}
We propose a learnable Color-aware Transform (CAT) that transforms images into illumination-invariant descriptors to facilitate stable enhancement.
Specifically, given an RGB image and two adjacent pixels $p_1, p_2$, we calculate the channel-wise color ratios by:

\begin{equation}\label{eq3}
   M_{r g}=\frac{R_{p_1} G_{p_2}}{R_{p_2} G_{p_1}},
    M_{r b}=\frac{R_{p_1} B_{p_2}}{R_{p_2} B_{p_1}},
     M_{g b}=\frac{G_{p_1}  B_{p_2}}{G_{p_2} B_{p_1}},
\end{equation}
For ease of exposition, we concentrate on $M_{r g}$ in the following discussion. Without loss of generality, all results derived for $M_{r g}$ will also hold for $M_{r b}$ and $M_{g b}$.
With the illumination assumption that $e^{C_{p_1}} \approx e^{C_{p_2}}$, the color ratio becomes independent of both the illumination intensity and direction and the change of viewpoint and object geometry, as shown by substituting Eq.~\eqref{eq2} into Eq~\eqref{eq3}:
\begin{equation}\label{eq4}
\begin{aligned}
   M_{r g}= & \frac{m^{p_1}(\vec{n}, \vec{l}) e^{R_{p_1}}(\lambda) \rho^{R_{p_1}}(\lambda) m^{p_2}(\vec{n}, \vec{l}) e^{G_{p_2}}(\lambda) \rho^{G_{p_2}}(\lambda)}{m^{p_2}(\vec{n}, \vec{l}) e^{R_{p_2}}(\lambda) \rho^{R_{p_2}}(\lambda) m^{p_1}(\vec{n}, \vec{l}) e^{G_{p_1}}(\lambda) \rho^{G_{p_1}}(\lambda)} \\
   = & \frac{\rho^{R_{p_1}} \rho^{G_{p_2}}}{\rho^{R_{p_2}} \rho^{G_{p_1}}} = \left(\frac{R}{G}\right)^{p_1} \left(\frac{R}{G}\right)^{p_2}.
\end{aligned}   
\end{equation}
It can be seen that dependencies on the surface normal and illumination direction $m(\vec{n}, \vec{l})$, and SPD $e$ are factored out. Therefore, this descriptor relies solely on the ratio of surface albedos, allowing it to possess illumination-invariant properties and be stable with SPD changes.

Given that extreme degradations can result in low values in dark regions and potential information loss,  we customize a color density parameter $S_{k}$ to adjust the color point density of the illumination-invariant feature as
\begin{equation}\label{eq5}
   S_k=\sqrt[k]{\sin(\frac{\pi I_{max}}{2})+\tau}, \\
   C_{map}=\{S_k M_{rg}, S_k M_{rb}, S_k M_{gb}\},
\end{equation}
where $I_{max}$ is the intensity map, calculated as $\mathbf{I}_{\max }=\max _{\mathbf{c} \in\{R, G, B\}}\left(\mathbf{I}_{\mathbf{c}}\right)$, $k \in \mathbb{Q}^{+}$ and $\tau = 1 \times 10^{-8}$.

To incorporate neighborhood relationships between pixels, we derive the resulting features by applying a local kernel $\mathcal{W}_{i}$ (\textit{e.g.}, Gaussian kernel or Laplacian kernel) to conduct convolution on the image $\textit{I}$, denoted as $f_{\mathcal{W}_{i}}(\textit{I})$:
\begin{equation}\label{eq6}
      f_{\mathcal{W}_{i}}(I)=\left[\begin{array}{l}
    \mathcal{W}_{i} \circledast (R/G),
    \mathcal{W}_{i} \circledast (R/B),
    \mathcal{W}_{i} \circledast (G/B)
    \end{array}\right].  
\end{equation}
Here, $\circledast$ denotes convolution.
However, the fixed formulation may not adequately capture the diverse and complex degradation scenarios, especially in some extremely dark conditions. Hence, we design a dynamic color-aware filter (DCAF) to enable adaptively learning of color-aware representation.

\textbf{Dynamic Color-aware Filter.} DCAF is designed to adaptively capture the color-aware representation of the illumination-invariant descriptor, conditioned on the input features.
First, we employ two convolution layers $f(x), h(x)$ to expand the channel dimension of the illumination-invariant feature $C_{map}$ from $3$ to $c$.
Next, we apply average pooling to  $f(C_{map})$ to generate a color-aware filter with kernel size $k$: $g_k(x) \in \mathbb{R}^{k \times k \times c}$.
At last, the feature map $h(C_{map})$ is convolved with the color-aware filter with depth-wise convolution to obtain a color-aware representation:
\begin{equation}\label{eq7}
    C^{'}_{map} =h(x) \otimes g_k(x),
\end{equation}

\subsection{Content-Noise Decomposition Network}
Low-light images usually suffer from serious noise disturbance and color distortion, which results in noisy illumination-invariant descriptors that are suboptimal for use as prior directly.
As such, the CNDN is specifically designed to explicitly estimate the noise distribution  and separately decompose content information and noise distribution.
We first employ several convolutional layers to obtain the embedded features from low-light image $x$ and noisy illumination-invariant descriptor $y$ as $x^{'}= Convs(x), y^{'}=Convs(y)$.
Next, we leverage a cross-attention~\cite{crossvit} layer to reinforce the content information in the low-light image as $x^{''}= CA(x^{'}, y^{'})$. Meanwhile, a self-attention layer~\cite{cai2023retinexformer} is used to further extract content information from the descriptor as $y^{''}= CA(y^{'})$.
The final content-enhanced low-light image and the noise distribution are expressed as $\Tilde{x}= Convs(x^{''} + y^{''}), n=Convs(y^{'} - y^{''})$, as shown in Fig.~\ref{fig:pipeline}.
The content-enhanced low-light image $\Tilde{x}$ is fed into downstream neural networks for further enhancement. In our work, we adopt a well-established image restoration framework, MST~\cite{cai2022mask}, as the enhancement network.

\section{Experiment}
\label{sec:exp}

\textbf{Datasets.}
We utilize five widely adopted LLIE benchmark datasets for evaluation, including LOL-v1~\cite{retinex_net}, LOL-v2-real~\cite{lol_v2}, LOL-v2-synthetic~\cite{lol_v2}, SID~\cite{sid}, and SMID~\cite{smid}.
During training, all images are randomly cropped into patches of size $256 \times 256$, with pixel values normalized to $\left[0,1\right]$.

\textbf{Implementation Details.}
Our model was implemented using Pytorch with a batch size of 16 on a signal Nvidia A100 GPU. It was trained for 400 epochs with the Adam optimizer ($\beta{1} = 0.9$, $\beta_{2} = 0.999$). During training, the learning rate is set to $2 \times 10^{-4}$ with a cosine annealing scheme.
More details of datasets and implementation, including objective functions, are provided in supplementary materials.

\subsection{Comparisons with State-of-the-art Algorithms}
We conducted comprehensive comparisons of our PiCat with 12 state-of-the-art LLIE methods, including RF~\cite{rf}, RetinexNet~\cite{retinex_net}, Sparse~\cite{lol_v2}, FIDE~\cite{fide}, DRBN~\cite{drbn}, Restormer~\cite{restormer}, MIRNet~\cite{mirnet}, SNR-Net~\cite{snr_net}, Retinexformer~\cite{cai2023retinexformer}, LLFormer~\cite{wang2023ultra}, HAIR~\cite{cao2024hair}, WaveMamba~\cite{zou2024wave}.
For evaluation metrics, we adopt PSNR and SSIM to evaluate the quantitative performance. Parameters (M) and FLOPs (G) are used for comparison of memory cost and complexity.

\begin{figure*}[h]

    \centering
  { 
      \includegraphics[width=0.98\linewidth]{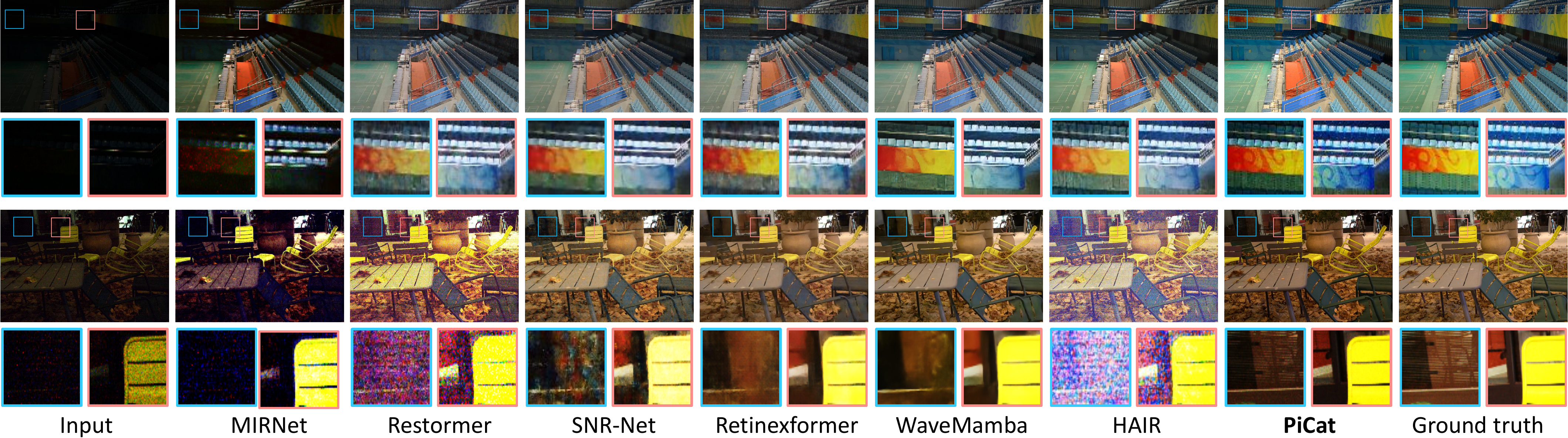}}
  \caption{Visual results on LOL-v1~\cite{retinex} (top) and SID~\cite{sid} (bottom). Brightness correction is equally applied to all cropped patches (\textcolor{SkyBlue}{blue box} and \textcolor{CarnationPink}{pink box}) for better detail comparison. Previous methods often fail due to noise, color distortion, or producing blurry and under- or over-exposed images. In contrast, our PiCat effectively removes noise and reconstructs well-exposed image details. (\textbf{Please Zoom in for the best view}.)} 
    \label{fig:visual_1} 
    \vspace{-5pt}
\end{figure*} 

\begin{table*}[t]
\hspace{-2.5mm}
\caption{We conduct extensive ablation studies of module effectiveness, model robustness, and generalization ability.}
	\label{tab:ablations}
	\subfloat[Break-down ablations. The baseline, CST, and DCAF denote MST, color space transform, and dynamic color-aware filter in CAT. \label{tab:breakdown}]{\vspace{1mm}
        \scalebox{0.83}{
        \begin{tabular}{c c c  c c c c}
            \toprule
            \rowcolor{color3} Baseline & CST & GDCAF & CNDN   & PSNR &SSIM \\
            \midrule
            \checkmark &          &          &          &24.87                 &0.843  \\
            \checkmark &\checkmark&          &          &26.15                 &0.857  \\
            \checkmark &\checkmark&\checkmark&          &26.61                 &0.871 \\
            \rowcolor{color4}\checkmark &\checkmark&\checkmark&\checkmark&\textcolor{red}{27.51}&\textcolor{red}{0.890} \\
            \bottomrule
	\end{tabular}}}\hspace{1mm}\vspace{0mm}
	\subfloat[Ablation of SPD pertubations. Gaussian noise with levels $\sigma \in \{15, 25, 50\}$ is injected to input to simulate SPD disturbance.\label{tab:spd}]{\vspace{1mm}
	\scalebox{0.83}{
		\begin{tabular}{l c c c c}
			\toprule
			\rowcolor{color3} Method &~$\sigma = 15$~ &~$\sigma = 25$~ &~$\sigma = 50$~ \\
			\midrule
			MIRNet    &21.49 &20.15 &18.71  \\
			MST  &22.05 &20.38 &19.06  \\
			  HAIR &21.31 &20.50 &19.25 \\
			\rowcolor{color4}PiCat &\textcolor{red}{26.74}&\textcolor{red}{26.18}&\textcolor{red}{25.81}\\
			\bottomrule
		\end{tabular}}}\hspace{1mm}\vspace{0mm}
        \subfloat[Ablation of the generalizability of PiCat. PiCat is integrated into Restormer~\cite{restormer} and SNR-Net~\cite{snr_net}, following the setting of PiCat + MST.\label{tab:generalizability}]{
		\vspace{1mm}\scalebox{0.83}{
			\begin{tabular}{l c c c c}
				\toprule
				\rowcolor{color3} Method  &Params (M) & FLOPs (G) &PSNR &SSIM\\
				\midrule
				Restormer &26.13 &144.25  &22.43 &0.823  \\
				\rowcolor{color4} Res-PiCat  &26.14 &144.61 &\textcolor{red}{24.71} &\textcolor{red}{0.840}  \\
				SNR-Net   &4.01 &26.35  &24.61 &0.842  \\
				\rowcolor{color4} SNR-PiCat  &4.02 &26.71 &\textcolor{red}{26.98} &\textcolor{red}{0.877} \\
				\bottomrule
	\end{tabular}}}\vspace{0mm}
\end{table*}

\textbf{Qualitative Results.}
Figure ~\ref{fig:visual_1} displays compelling visual comparisons between our PiCat method and SOTA methods. 
For a thorough examination of the results, zooming in on specific areas is recommended.
As observed, all previous methods struggle to capture intricate texture details and frequently yield artifacts. Notably, the painted walls in the top row (highlighted in the blue box on the left patch) show significant failures in detail restoration.
Moreover, existing methods either overlook noise suppression or fail to address underexposed and overexposed areas, as showcased in the bottom row.
In contrast, our PiCat method excels in enhancing visibility and texture restoration, robustly suppressing noise, and preserving natural color.

\textbf{Quantitative Results.}
We provide the quantitative comparison in Table~\ref{tab:quantitative}. Given that PiCat is exceptionally lightweight, we have developed an expanded version, PiCat-large, by stacking additional attention layers in MST.
As can be seen, our PiCat outperforms existing methods on all five datasets while requiring the least computational and memory costs.
Especially when compared with the recent method WaveMamba, our PiCat-large achieves 0.97, 0.27, 0.96, and 0.75 dB improvements on LOL-v1, LOL-v2-real, LOL-v2-synthetic, and SID datasets. Meanwhile, our PiCat-large only costs $55.6\%$ (13.67/24.50) FLOPs and $95.1\%$ (0.97/1.02) parameters.

To further illustrate the performance edge of our PiCat method over SOTA approaches, we provide a performance comparison map in Fig. ~\ref{fig:performance}. Our PiCat is prominently positioned in the upper left corner, indicating the best performance in PSNR combined with low temporal complexity.
Additionally, the compact circular representation underscores its minimal spatial complexity, making PiCat a highly efficient and effective solution for low-light image enhancement.
Finally, we provide the inference time comparison of the best batch of methods in Table \ref{tab:table}, which shows that our method is the fastest model and demonstrates great potential in the application of real-time scenarios.


\begin{table}[t]
\centering
\caption{Inference time comparison using an input size of $600 \times 400$.}
\label{tab:table}
    \resizebox{0.48\textwidth}{!}{
    \large
    \begin{tabular}{*{7}{c}}
        \toprule
       Method & MIRNet &  SNR-Net & RetinexFormer &  HAIR &  WaveMamba &  Ours \\
        \midrule
        Speed(s) & 0.246 & 0.123 & \textcolor{blue}{0.094} & 0.205 & 0.118 & \textcolor{red}{0.087} \\
        
        \bottomrule
    \end{tabular}
    }
\end{table}

\subsection{Ablation Study}

\begin{figure}[t]

    \centering
  { 
      \includegraphics[width=0.95\linewidth]{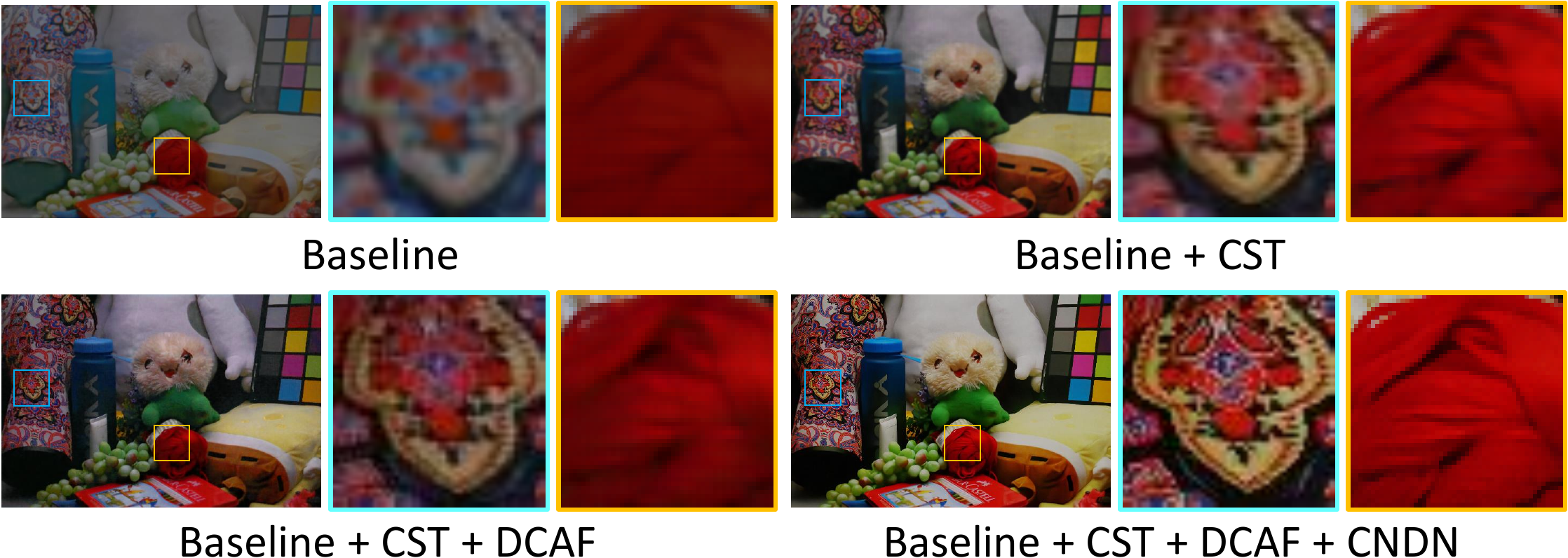}}
  \caption{Visual results of break-down ablations. The last candidate is the full implementation of our PiCat.} 
    \label{fig:visual_2} 
    \vspace{-5pt}
\end{figure}

\textbf{\textit{Is PiCat really effective?}}
To evaluate the effectiveness of our proposed PiCat, we conducted a breakdown of the ablation studies for the CAT and CNDN components to analyze the impact of each part. As shown in Table~\ref{tab:breakdown}, the baseline for comparison is the pure MST structure proposed by \cite{cai2022mask}. CST and GDCAF refer to the color space transform and dynamic color-aware filter, respectively. The results indicate that the color space transform contributes the most to performance improvement, yielding a 1.28 dB increase in PSNR. This highlights the effectiveness of the illumination-invariance transformation. Furthermore, the model achieves optimal performance only when both GDCAF and CNDN are included.
Additionally, we provide visual comparisons in Fig.~\ref{fig:visual_2}. The baseline model fails to produce vibrant colors, resulting in blurred images and color distortion. In contrast, the complete model (shown in the bottom right corner) successfully predicts vivid colors and maintains clean textures.

\textbf{\textit{Is PiCat really robust to SPD perturbations?}}
To investigate the robustness of our proposed PiCat against SPD perturbations, as stated in Section~\ref{sec:intro}, we injected Gaussian noise at various levels, specifically $\sigma \in \{15, 25, 50\}$, into the training samples to simulate SPD disturbances. As shown in Table~\ref{tab:spd}, other models such as MIRNet~\cite{mirnet}, MST~\cite{cai2022mask}, and HAIR~\cite{cao2024hair} experienced significant performance degradation with noisy input. For example, with $\sigma = 15$, MIRNet exhibited a degradation of 10. 98\%, MST showed a decrease of 11. 34\%, and HAIR had a drop of 7. 83\% in PSNR. In contrast, our PiCat demonstrated a much smaller degradation of only 2.80\% in PSNR under the same conditions. This indicates that our PiCat is significantly more robust to SPD perturbations.

\textbf{\textit{Can PiCat serve as a plug-and-play module to enhance the performance of downstream frameworks?}}
To investigate whether PiCat is a generalized enhancement module for different image restoration frameworks, we plug PiCat as a pre-processing module into different image restoration frameworks, Restormer~\cite{restormer} and SNR-Net~\cite{snr_net}, following the same setting as our PiCat + MST~\cite{cai2022mask} for fair comparison. As can be seen in Table~\ref{tab:generalizability}, embedded with PiCat improves the performance of Restormer for 2.28 dB and SNR-Net for 2.37 dB in PSNR, respectively, indicating that our PiCat can serve as a plug-and-play module for existing image restoration frameworks.

\section{Conclusion}
In this paper, we presented PiCat, a lightweight and effective color transformation designed for stable and high-quality low-light image enhancement. PiCat has two main components: a Color-aware Transform (CAT) and a Content-Noise Decomposition Network (CNDN). 
The CAT extracts physics-informed priors based on the Lambertian assumption, transforming low-light observations into illumination-invariant color descriptors. Subsequently, the CNDN complements this by explicitly estimating the noise distribution and separately decomposing content representation from the estimated noise, enabling the transfer of noise-free content information into low-light features.
Comprehensive experiments on five low-light image enhancement benchmarks demonstrate the effectiveness and robustness of our proposed PiCat.

\bibliographystyle{IEEEbib}
\bibliography{myRefs}

\end{document}